\pdfoutput=1
\documentclass[11pt]{article}

\usepackage{acl}
\usepackage{amsmath,bm}
\usepackage{enumitem}
\usepackage{subcaption}
\usepackage{graphicx}
\usepackage{multirow}
\newtheorem{definition}{Definition}
\usepackage{times}
\usepackage{latexsym}

\usepackage[T1]{fontenc}
\usepackage[utf8]{inputenc}

\usepackage{microtype}

\title{Towards Job-Transition-Tag Graph for a Better Job Title Representation Learning}
\author{Jun Zhu, Céline Hudelot \\
Laboratoire Mathématiques et Informatique pour la Complexité et les Systèmes\\ 
CentraleSupélec, Université Paris-Saclay \\
Gif-sur-Yvette, France
\\ \texttt{\{jun.zhu, celine.hudelot\}@centralesupelec.fr}}
\begin{document}
\maketitle
\begin{abstract}
%The job title is a short term, but it usually carries essential information. Learning the appropriate representation of job titles can benefit various downstream tasks, such as modeling career profiles to provide career guidance and categorizing jobs to improve the search and recommendation. 
Works on learning job title representation are mainly based on \textit{Job-Transition Graph}, built from the working history of talents. However, since these records are usually messy, this graph is very sparse, which affects the quality of the learned representation and hinders further analysis. To address this specific issue, we propose to enrich the graph with additional nodes that improve the quality of job title representation. Specifically, we construct \textit{Job-Transition-Tag Graph}, a heterogeneous graph containing two types of nodes, i.e., job titles and tags (i.e., words related to job responsibilities or functionalities). Along this line, we reformulate job title representation learning as the task of learning node embedding on the \textit{Job-Transition-Tag Graph}. Experiments on two datasets show the interest of our approach.
\end{abstract}

\section{Introduction}
\label{sec: Introduction}
%The learning of job title representations has received much attention in the recruitment field. The learned representations are used for various tasks, e.g. These works use different learning methods, but they are all based on the construction of \textit{Job-Transition Graph} $\mathcal{G}^{jj}$. 

The learning of job title (referred to as job for short)~\footnote{In this paper, we use \textit{job title} and \textit{job} interchangeably.} representation has received much attention in the recruitment field because the learned representation is beneficial to various tasks, such as job recommendation~\citep{dave2018combined,liu2019tripartite}, job title benchmarking~\citep{zhang2019job2vec}, and job mobility prediction~\citep{ zhang2021attentive,zhu2021improving}. However, in practice, learning a good representation is challenging for the following reasons: (i) \textbf{Noisy data}: job title data is noisy due to personal subjective reasons (i.e., spelling errors) or objective reasons (i.e., resume parsers are not perfect). (ii) \textbf{Messy data}: job titles are messy because people have different ways of thinking, and naming conventions vary by company and industry. For example, there are many alternative job titles for the same position, e.g., ``purchasing clerk'' and ``buyer''. Another problem is that due to the ambiguity of certain terms, they can refer to different positions in different contexts, e.g., ``registered nurses sandwich rehab'' and ``sandwich maker''. For these reasons, standard semantic-based approaches that aggregate (e.g., mean or sum) word representations to get job title semantic representation may lead to mismatches. Moreover, these methods ignore hidden relationships between job titles, e.g., titles in the same resume may be similar. ~\citep{dave2018combined,zhang2019job2vec} learn representations from graphs. They create graphs from career trajectories, where nodes represent job titles and edges represent job transitions. Then they design different loss functions to embed the nodes into a low-dimensional space. However, the generated graphs are usually sparse due to the above reasons, limiting the performance of graph-based methods. Standardizing job titles before generating graphs can alleviate the sparsity issue to a certain extent, but at the cost of losing some information. To tackle these challenges, we propose to enrich graphs with structured contextual information and learn job title representations through network embedding methods. More specifically, inspired by domain-specific Named Entity tags (i.e., \textit{RES}ponsibility and \textit{FUN}ction) proposed in~\citep{liu2019ipod}, we treat the job title as a combination of responsibilities, functionalities, and other additional information.
 % The \textit{responsibility} part describes the responsibility-level of the position, e.g., senior, director and manager, the \textit{functionality} part describes the core function/operation type, e.g., marketing, security and education, and the \textit{additional information} contains some personal-specific information, such as company name and geographic location. 
Words related to \textit{responsibility} and \textit{functionality} are defined as tags. We assume that job titles with the same tag describe similar job responsibilities or functionalities, making them more likely to have similar representations. Along this line, we construct \textit{Job-Transition-Tag Graph}, a heterogeneous graph containing two types of nodes, i.e., job titles and tags, which carries more information, thereby alleviating the sparsity problem.

%\vspace{-0.02cm}
 %% ---------Sec: Methodology -------.
\section{Methodology}
\label{sec: Methodology}
%This section explains the inspiration for constructing \textit{Job-Transition-Tag Graph}.

%This section introduces previous explicit and implicit job title representation learning studies based on \textit{Job-Transition Graph}, and explain the inspiration for constructing \textit{Job-Transition-Tag Graph}.
%Throughout this paper, we use uppercase calligraphic characters to denote sets (e.g., $\mathcal{J}$), the cardinality of a set is denoted by vertical bars $\vert \cdot \vert$, and non-bold letters (e.g., $n$, $N$) to represent scalars. Following the notations, 

\subsection{Preliminaries}
A graph/network is represented as  $\mathcal{G}=(\mathcal{V},\mathcal{E})$, with node set $\mathcal{V}$ and edge set $\mathcal{E}$%~\footnote{In this paper, we use \textit{graph} and \textit{network} interchangeably.}
. Nodes and edges can optionally have a type, so a graph can be homogeneous or heterogeneous. In the recruitment field, the career trajectory of talents can be represented by graphs. Formally, consider a job seeker set $\mathcal{U}$ and their working history set $\mathcal{H}=\{H^{u}\}_{u\in \mathcal{U}}$, where the working history of each $u$ is represented as a sequence of $n$ work records ordered by time $H^{u} = \{J_{1}, \dots, J_{n}\}$. The $i$-th record $J_{i}$ is denoted by $(j_{i}, p_{i})$, indicating that $u$ is engaged in a position (titled $j_{i}$) during the $p_{i}$ period. The set of job titles $j_{i}$ that occurred in $\mathcal{H}$ is denoted as $\mathcal{J}$. Based on $\mathcal{H}$, \textit{Job-Transition Graph} (Figure~\ref{fig:Job-Transition}) can be constructed, which is formally defined as:

\begin{definition}[\textbf{Job-Transition Graph}] is defined as a directed homogeneous graph $\mathcal{G}^{jj}=(\mathcal{J}, \mathcal{E}^{jj})$ generated from $\mathcal{H}$, where $\mathcal{J}$ is a set of job titles, and the edge $e^{jj}_{xy}\in\mathcal{E}^{jj}$ represents the job transition from the former job $j_{x}$ to the next job $j_{y}$.
\end{definition}

%To be noted, in our setting, \textit{Job-Transition Graph} is a homogeneous graph, which only has one node type (i.e., job title) and one edge type (i.e., job transition). 
%An example is given in Figure~\ref{fig:Job-Transition}, where the blue circles represent job titles, and black directed lines represent job transitions between job titles. 

\subsubsection{Learning from Job-Transition Graph: An Overview}
$\mathcal{G}^{jj}$ is often used for job title representation learning tasks. The current procedure is first to build a $\mathcal{G}^{jj}$, and then learn job title representation from it. More specifically, ~\citep{dave2018combined} first builds $\mathcal{G}^{jj}$ and the other two graphs. Then, the Bayesian personalized ranking and margin-based loss functions are used to learn job title representations from graphs. Job2Vec~\citep{zhang2019job2vec} constructs a $\mathcal{G}^{jj}$, where the node denotes a job title affiliated with the specific company, and a multi-view representation learning method is proposed. ~\citep{zhang2021attentive} adds company nodes in $\mathcal{G}^{jj}$ to build a heterogeneous graph. Then they use a graph neural network to represent the nodes. As mentioned in Section~\ref{sec: Introduction}, the job title and job transition data are messy. Therefore, $\mathcal{G}^{jj}$ may be sparse~\citep{zhang2019job2vec}, which we will further prove in Section~\ref{sec: Datasets}. In order to alleviate this issue, a simple method is to standardize job titles and then construct a normalized and denser graph based on the standardized job titles. For example, ~\citep{dave2018combined} normalizes titles by using Carotene~\citep{javed2015carotene}, ~\citep{zhang2019job2vec} aggregates titles by filtering out low frequency words, and~\citep{zhang2021attentive} unifies titles according to IPOD~\citep{liu2019ipod}. However, the standardization of job titles may lose some specific information. Furthermore, these methods either ignore the semantic information contained in job titles~\citep{dave2018combined,zhang2021attentive} or separate the semantic information from the graph topology~\citep{zhang2019job2vec}.

%improving talent profile modeling and facilitating search and recommendation. The learning of job title representations has received much attention in the recruitment field. The learned representations are used for various tasks, e.g. These works use different learning methods, but they are all based on the construction of \textit{Job-Transition Graph} $\mathcal{G}^{jj}$. 

\subsubsection{Job Title Composition}
\label{subsec: Job Title Composition}
Generally, a job title consists of three parts~\cite{liu2019ipod,zhang2019job2vec}: (i) \textbf{Responsibility}: describes the role and responsibility of a position from different levels (e.g., director, assistant, and engineer). (ii) \textbf{Functionality}: describes the business function of a position from various dimensions (e.g., sales, national and security). (iii) \textbf{Additional Information}: contains personal-specific information. We denote the words related to \textit{responsibility} and \textit{functionality} as tags, and they form a tag set $\mathcal{T}$. These tags are the essence of the job title and provide important information about the position. However, few works directly include this information in the representation learning scheme. In this paper, we consider these tags when generating graphs. These tags can alleviate the graph sparsity problem of $\mathcal{G}^{jj}$ and provide additional information for the learning of job title representations.

\begin{figure*}[ht]
    \centering
    \begin{subfigure}[b]{0.5\textwidth}
    \centering
    \includegraphics[width=1.\textwidth]{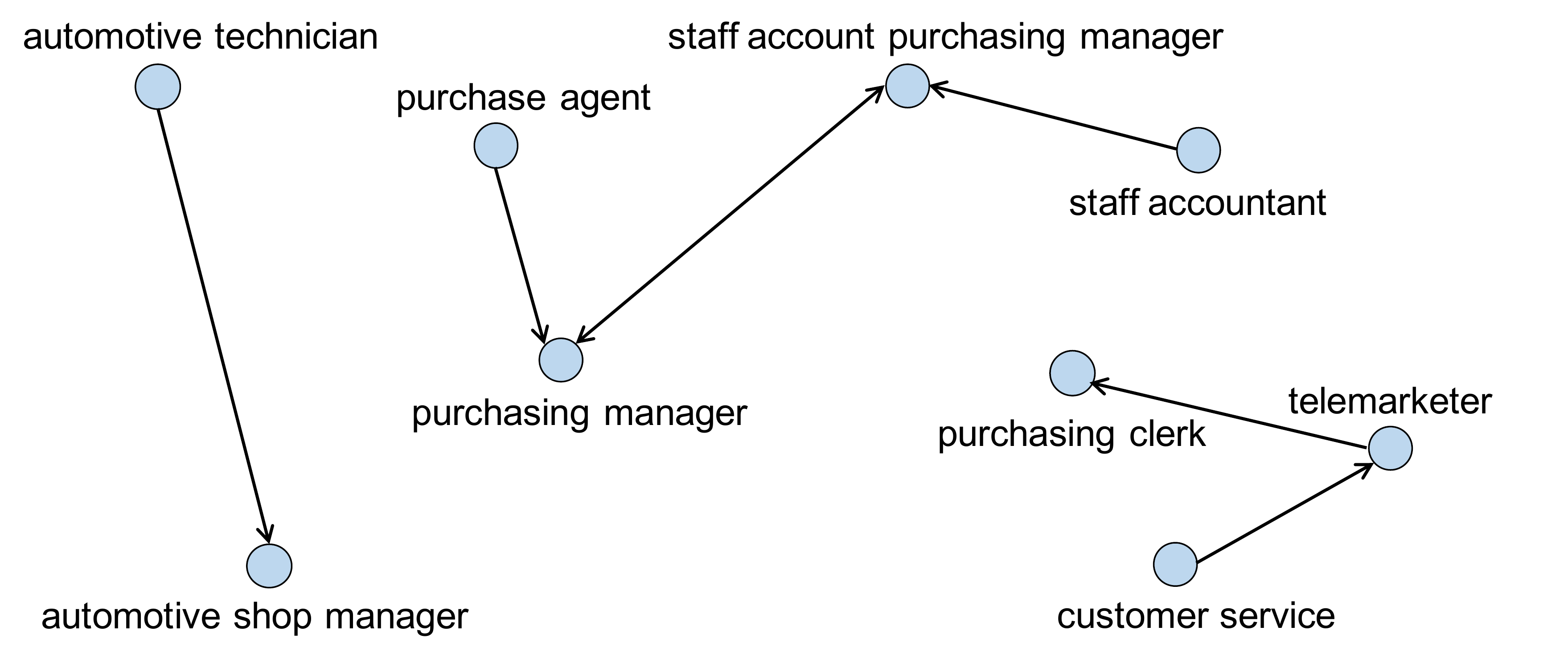}
    \caption{Job-Transition graph.}
    \label{fig:Job-Transition}
    \end{subfigure}
    \hspace{-0.5em}
    \begin{subfigure}[b]{0.5\textwidth}
    \centering
    \includegraphics[width=1.\textwidth]{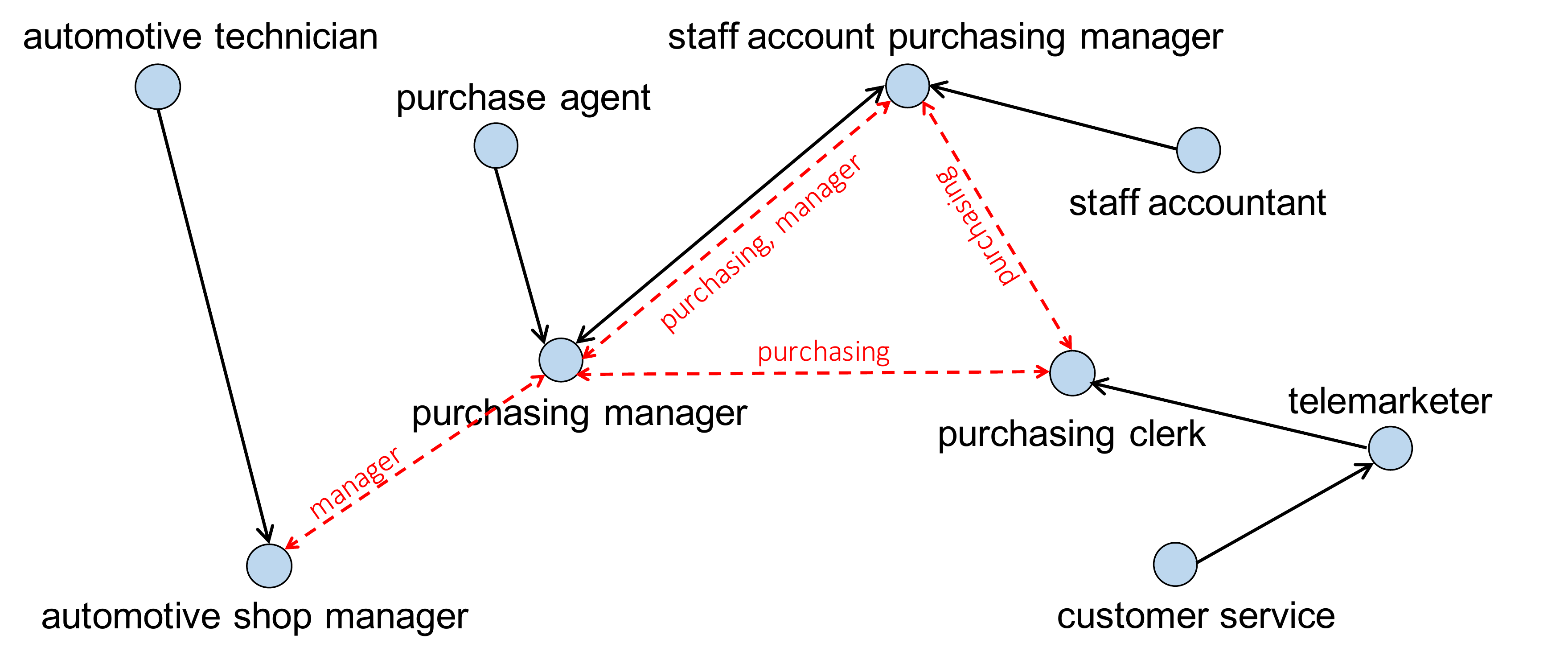}
    \caption{Enhanced Job-Transition graph.}
    \label{fig:E_Job-Transition}
  \end{subfigure}
  \hspace{-0.5em}
  \begin{subfigure}[b]{0.5\textwidth}
    \centering
    \includegraphics[width=1\textwidth]{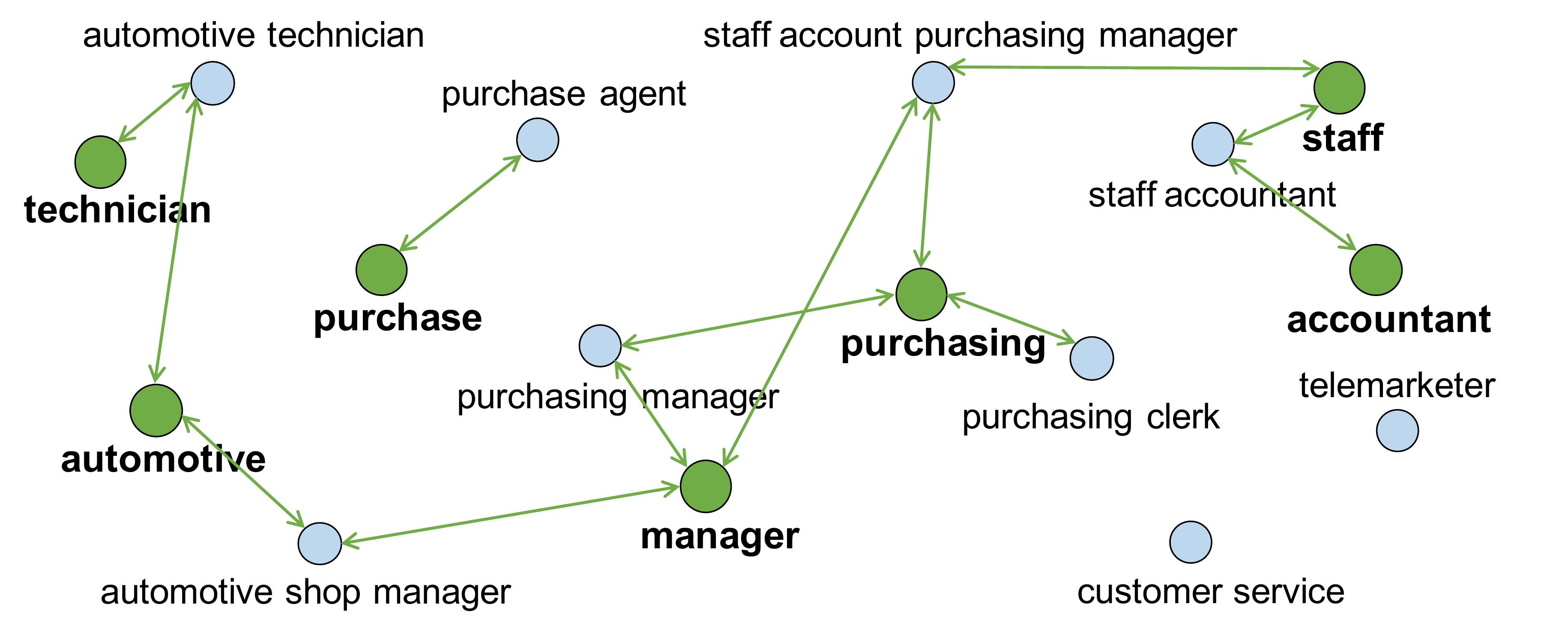}
    \caption{Job-Tag graph.}
    \label{fig:Job-Tag}
  \end{subfigure}
   \hspace{-0.5em}
  \begin{subfigure}[b]{0.5\textwidth}
    \centering
    \includegraphics[width=1\textwidth]{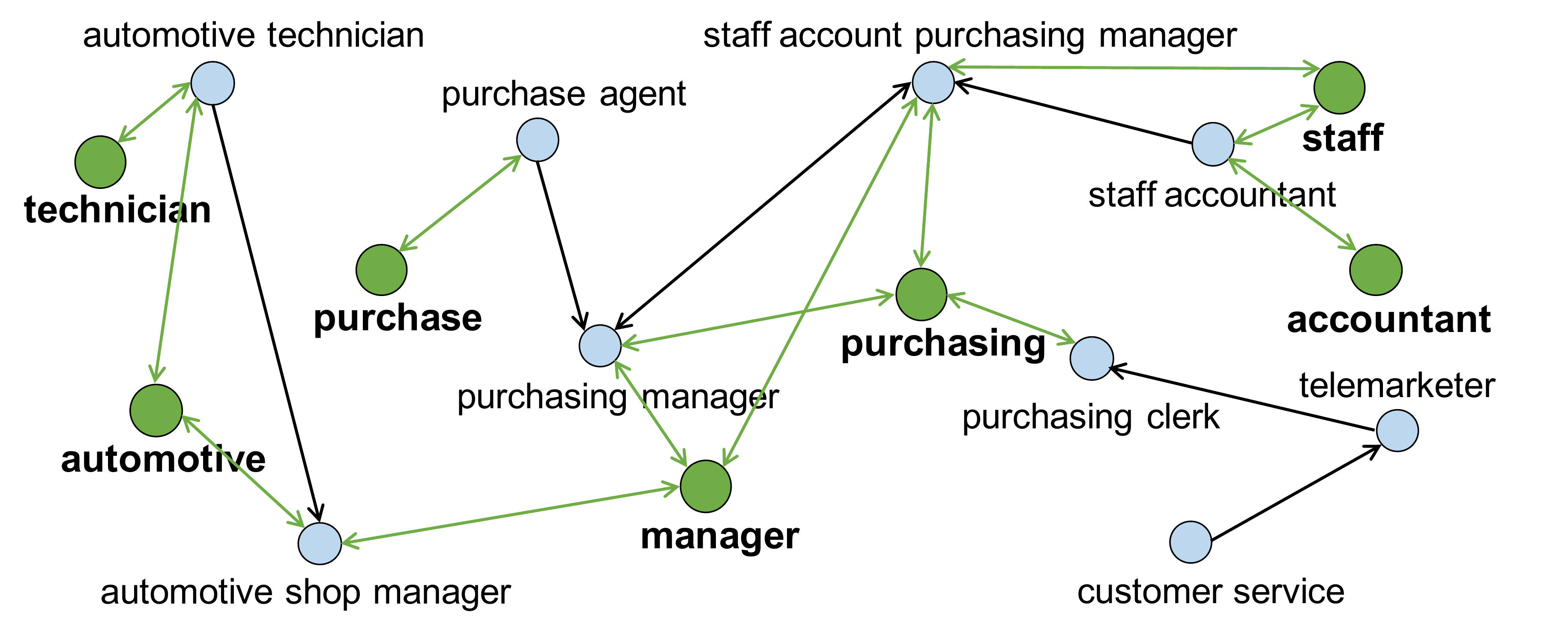}
    \caption{Job-Transition-Tag graph.}
    \label{fig:Job-Transition-Tag}
  \end{subfigure}
  \caption{Examples of four types of graphs with blue/green circles representing job titles/tags. Black lines represent job transitions, red dotted lines represent enhanced edges, and green lines represent ``has/in'' relationships. }
\end{figure*}

\subsection{Proposed Graphs}
\label{subsec: Problem Formulation}
In order to address the sparsity issue of $\mathcal{G}^{jj}$, we consider adding more information when generating graphs, i.e., tags related to job responsibilities or functionalities, driven by the job title composition. Along this line, we define various types of graphs:
\begin{definition}[\textbf{Enhanced Job-Transition Graph}] is based on $\mathcal{G}^{jj}$ with additional enhanced edges. It is defined as $\mathcal{G}^{jj}_{E}=(\mathcal{J}, \mathcal{E}^{jj} \cup \mathcal{E}^{jj}_{E})$, where $\mathcal{E}^{jj}_{E}$ is a set of enhanced edges. More specifically, if $j_{x}$ and $j_{y}$ share a tag $w$, then we add a bi-directional edge between them, i.e., $e^{jj}_{xy}$ and $e^{jj}_{yx}$. %The word $w$ can belong to $\mathcal{T}$ or other predefined vocabulary sets.
\end{definition}

As shown in Figure ~\ref{fig:E_Job-Transition}, red dashed lines represent additional enhanced edges, e.g., ``purchasing manager'' shares the tag ``purchasing'' with ``purchasing clerk'', so we add edges between them.

% Note that the \textit{Enhanced Job-Transition Graph} can be either homogeneous or heterogeneous, depending on how we define the edge type of the red dashed line. If the edge type is expressed as ``previous job'', similar to the \textit{Job-Transition Graph}, the \textit{Enhanced Job-Transition Graph} is homogeneous. While if we define the edge type as ``share a common word'', it is heterogeneous, this is analogous to \textit{Job-Transition-Tag Graph}, which is defined in Definition~\ref{def:Job-Transition-Tag Graph}. In this work, we consider that \textit{Enhanced Job-Transition Graph} is homogeneous.

\begin{definition}[\textbf{Job-Tag Graph}] is defined as a heterogeneous graph $\mathcal{G}^{jt}=(\mathcal{J}\cup\mathcal{T}, \mathcal{E}^{jt})$, with job titles and tags, two node types. $\mathcal{E}^{jt}$ is a set of bi-directional edges between a job title and a tag, representing the ``has/in'' relationship.
\end{definition}

An example of the ``has/in'' relationship is given in Figure~\ref{fig:Job-Tag} (i.e., the green line), where the job title ``automotive technician'' \textit{has} a tag ``automotive'', and ``automotive'' is \textit{in} ``automotive technician''. In order to aggregate more information, we further combine \textit{Job-Transition Graph} and \textit{Job-Tag Graph} to build \textit{Job-Transition-Tag Graph}:
\begin{definition}[\textbf{Job-Transition-Tag Graph}]
\label{def:Job-Transition-Tag Graph} is defined as a heterogeneous graph $\mathcal{G}^{jtj}=(\mathcal{J}\cup\mathcal{T}, \mathcal{E}^{jj}\cup\mathcal{E}^{jt})$ with job titles and tags, two node types, and transition and  ``has/in'', two edge types.
\end{definition}

Inspired by the achievements of network embedding models in the node representation learning problem~\citep{hamilton2017representation}, we apply different network embedding models to learn job title representation from the graphs defined above.

%We formally define the \textit{Job Title Representation Learning} problem as follows:\textbf{Given}: a set of working histories $\mathcal{H}$ and an optional tag set $\mathcal{T}$~\footnote{Tag set can be any pre-defined set or a vocabulary set customized from $\mathcal{H}$.} to construct various types of graphs.\textbf{Obtain}: job title representations learned from graphs by using network embedding methods. 

%% ---------Sec: Experiments -------.
\section{Experiments}
\label{sec: Experiments}
We evaluate the proposed job title representation learning scheme through (i) a node classification task (i.e., \textit{job title classification}) and (ii) a link prediction task (i.e., \textit{next-job prediction}). This section will first describe the two datasets used and experimental settings, followed by discussing the results.

%% ---------Subsec: Datasets -------.
\subsection{Datasets}
\label{sec: Datasets}
\textbf{CareerBuilder12 (CB12):} an open dataset from a Kaggle competition.~\footnote{\url{https://www.kaggle.com/c/job-recommendation}}
%~\footnote{\url{https://www.careerbuilder.com/}}.
 It contains a collection of working experiences represented by sequences of job titles. For the node classification task, we use \textit{AutoCoder}~\footnote{\url{http://www.onetsocautocoder.com/plus/onetmatch}} to assign a SOC 2018 %~\footnote{\url{https://www.bls.gov/soc/}} 
 to each job title. The labeling details are given in Appendix~\ref{app: Job Title Label Assignment}. \textbf{Randstad:} a private French resume dataset provided by Randstad company,
where each resume is parsed into multiple sections, e.g., \textit{PersonalInformation}, \textit{EducationHistory} and \textit{EmploymentHistory}. An example is given in Figure~\ref{fig:parsed_resume} of Appendix. Graphs are built from \textit{EmploymentHistory} section. 

Both datasets use a tree-like taxonomy, as described in Appendix~\ref{app: Job Title Label Assignment} and ~\ref{app: Randstad Data Description}. For example, from root class to leaf class, the \textit{CB12} taxonomy is organized as $\textit{MajorGroup}\rightarrow \textit{MinorGroup} \rightarrow \textit{BroadGroup} \rightarrow \textit{DetailedOccupation}$. Consistent with the reality of the recruitment market, some occupations rarely appear in both datasets. To balance the data, we filter out occupations with fewer than 200 occurrences, i.e., \textit{MinorGroup} for \textit{CB12} and \textit{JobGroup} for \textit{Randstad}. Also, for graph construction, we remove working histories with less than two work records, i.e., $\vert H^{u}\vert < 2$.

\begin{table}[th]
    \centering
    \small
    \begin{tabular}{p{1.1cm}|p{0.15cm}p{0.5cm}p{0.6cm}p{0.6cm}p{1.05cm}p{0.6cm}}
    \hline
    & $\#$C & $\#$W &  $\vert\mathcal{J}\vert$ &  $\vert\mathcal{E}^{jj}\vert$ & $\vert\mathcal{E}^{jj}_{E}\vert$ & $\vert\mathcal{E}^{jt}\vert$\\ 
    \hline
    \textbf{CB12} & 16 & 1,682 & 9,216 & 20,640 & 6,477,819 & 22,149\\
    \textbf{Randstad} & 18 & 2,303 & 12,864 & 36,722 & 6,663,267 & 22,897\\ 
    \hline
    \end{tabular}
    \caption{Statistics of datasets and corresponding graphs, \#C is the number of categories, and \#W represents the vocabulary size for node one-hot encoding.}
\label{tab:Statistics}
\end{table}
%\vspace{-0.5cm}

\begin{table*}[th]
    \footnotesize
    \centering
    \begin{tabular}{p{0.05cm}|p{0.4cm}|p{1.3cm}p{1.3cm}p{1.5cm}|p{1.3cm}p{1.3cm}p{1.5cm}|p{1.3cm}p{1.3cm}}
    \hline
    && N2V & GCN & GAT & M2V & RGCN & HAN & W2V & BERT \\
    \hline
    \multirow{4}{*}{\rotatebox{90}{\textbf{CB12}}} & $\mathcal{G}^{jj}$  & 0.206/0.360 & 0.576/0.688 & 0.568/0.664 & 0.154/0.334 & 0.524/0.637 & 0.670/0.747 &  \multirow{4}{*}{0.713/0.767}& \multirow{4}{*}{0.688/0.719}\\
    & $\mathcal{G}^{jj}_{E}$ & \underline{0.599/0.714} & \underline{0.628/0.720} & \underline{0.692/0.759} & 0.571/0.688 & 0.591/0.701 & 0.698/0.781\\
    & $\mathcal{G}^{jt}$ & - & - & - & \underline{0.588/0.692} &   0.687/0.752 & 0.703/0.766\\
    & $\mathcal{G}^{jtj}$ & - & - & - &  \underline{0.588/0.692} & \underline{0.703/0.766} & \textbf{\underline{0.742/0.797}}\\
    \hline
    \hline
    \multirow{4}{*}{\rotatebox{90}{\textbf{Randstad}}} & $\mathcal{G}^{jj}$  & 0.201/0.304 &  \underline{0.520}/0.616 & 0.529/0.593 & 0.166/0.282 & 0.388/0.536 & 0.592/0.665 & \multirow{4}{*}{0.595/0.671} & \multirow{4}{*}{0.580/0.609}\\
     & $\mathcal{G}^{jj}_{E}$ & \underline{0.523/0.623} & 0.484/\underline{0.621} & \underline{0.607/0.677} & 0.469/0.585 & 0.452/0.580 & 0.607/0.689\\
     & $\mathcal{G}^{jt}$ & - & - & - & \underline{0.590/0.665}  & 0.552/0.643 & 0.572/0.663 & \\
    & $\mathcal{G}^{jtj}$ & - & - & - & \underline{0.590/0.665} & \underline{0.600/0.678} & \textbf{\underline{0.641/0.708}}\\
    \hline
    \end{tabular}
    \caption{Job title classification results (Macro-F1/Micro-F1). The score in bold is the best among all methods applied to all graphs, and the scores underlined are the best in all graphs of each method. For \textit{M2V}, we report the best results obtained by the meta-path \textit{Job-Tag-Job}.}
    \label{tab:classification}
\end{table*}

In order to generate tags, we first tokenize titles into tokens and remove stopwords, numbers, and punctuation. Then, we use the Top-200 tokens that appear most frequently in job titles and belong to IPOD~\citep{liu2019ipod} as tags. The details of tag generation are given in Appendix~\ref{app: Tags Generation}. We assign the one-hot encoding of the corresponding title for each title node as the node feature. The vocabulary set is obtained by filtering words with a frequency of 1 from the tokenized job titles. Statistics for datasets and graphs are summarized in Table~\ref{tab:Statistics}. We can observe that $\mathcal{G}^{jj}$ (i.e., $\vert\mathcal{J}\vert$ and $\vert\mathcal{E}^{jj}\vert$) are sparse.

%% ---------Subsec: Experimental Settings -------.
\subsection{Experimental Settings}

\noindent\textbf{For \textit{job title classification} task:} we classify job titles into root categories in this work, i.e., \textit{MajorGroup} for \textit{CB12} and \textit{JobClass} for \textit{Randstad}. We randomly split the data into training/validation/test sets with a ratio of 60\%/20\%/20\%.

\noindent\textbf{For \textit{next-job prediction} task:} we treat it as a link prediction task on \textit{Job-Transition Graph} to predict whether there exists an edge (transition) between two nodes (job titles). We keep the same split ratio on positive/negative edges, where negative edges are randomly picked from unconnected node pairs (i.e., the same size as positive edges). 
 
We evaluate the performance of our proposed learning scheme against the following baselines. A detailed description of these baselines is provided in Appendix~\ref{app: Baseline Description}.

\begin{itemize}
[noitemsep,topsep=1mm,leftmargin=15pt]
%[topsep=1mm,leftmargin=15pt]

\item \textbf{Homogeneous}: \textit{Node2Vec (N2V)}~\citep{grover2016node2vec}, \textit{GCN}~\citep{kipf2016semi} and \textit{GAT}~\cite{velivckovic2017graph}.
\item \textbf{Heterogeneous}: \textit{Metapath2Vec (M2V)}~\cite{dong2017metapath2vec}, \textit{RGCN}~\cite{schlichtkrull2018modeling} and \textit{HAN}~\cite{wang2019heterogeneous}.
\item \textbf{Semantic-based}: \textit{Word2Vec (W2V)}~\citep{le2014distributed} and \textit{BERT}~\citep{devlin2018bert}.
\end{itemize}
Our implementation is based on the DGL package ~\citep{wang2019deep}. ~\footnote{Source code will be available at \url{https://github.com/zhujun81/Job_title_representation}.} In both tasks, for unsupervised methods, node representations are learned from the entire dataset. Then train the logistic regression classifier on both the training and validation sets. Each semi-supervised model is trained on the training set, and the parameters are optimized on the validation set. The final performance is evaluated on the test set. To ensure fairness, we keep the same data split for both methods, repeat each prediction experiment ten times, and report the average performance scores (i.e., Macro-F1 and Micro-F1 for \textit{job title classification} and AUC for \textit{next-job prediction}). For details of other parameter settings, see Appendix ~\ref{app: Parameter Settings}.

\subsection{Results}
\label{sec: Results}

\subsubsection{Job Title Classification}
Table~\ref{tab:classification} summarizes the best results of all methods on different graphs. We have the following observations: (i) Among all graphs, all models generally have the lowest scores on $\mathcal{G}^{jj}$ because this graph is often sparse and can only provide limited information. (ii) All models perform better on $\mathcal{G}^{jj}_{E}$ (except Macro-F1 of \textit{GCN}) than $\mathcal{G}^{jj}$, which shows that the enhanced edges provide additional information. One interpretation for enhanced edges is adding semantic information, i.e., two job titles are more likely to be similar if they share the same word, represented by edges from the graph perspective. (iii) The heterogeneous models perform well on our proposed $\mathcal{G}^{jtj}$, which indicates that the added tag nodes can effectively improve the quality of representation. Note that we did not apply homogeneous methods to $\mathcal{G}^{jtj}$, but the results on $\mathcal{G}^{jj}_{E}$ prove that the information given by tags is useful. (iv) The models with attention mechanisms outperform the models without attention, demonstrating that the attention mechanism is good at capturing important information from noisy graphs.

\begin{table*}[th]
    \footnotesize
    \centering
    \begin{tabular}{p{0.4cm}|ccccccccc}
    \hline
    & N2V & GAN & M2V & HAN (Dot) & W2V (Dot) & BERT (Dot) & W2V (Hadamard) & BERT (Hadamard) \\
    \hline
    $\mathcal{G}^{jj}$ & 0.564 & 0.704 & 0.548 & 0.685 & \multirow{4}{*}{0.763} & \multirow{4}{*}{0.477} & \multirow{4}{*}{0.777} & \multirow{4}{*}{\textbf{0.840}}\\
    $\mathcal{G}^{jj}_{E}$ & 0.692 & 0.789 & 0.593 & 0.792\\
    $\mathcal{G}^{jt}$ & - & - & 0.604 & 0.768\\
    $\mathcal{G}^{jtj}$ & - & - & 0.604 & \underline{0.833}\\
    \hline
    \end{tabular}
    \caption{Next-job prediction results (AUC) on \textit{CB12}. The bold score is the best among all methods, and the underlined score is the second-best.}
    \label{tab:prediction}
\end{table*}
\subsubsection{Next-Job Prediction}
We further evaluate the learning scheme using \textit{next-job prediction}, which can be viewed as a link prediction task to predict whether a position will be recommended as the next-job. For unsupervised methods, edge features are represented by applying binary operators~\citep{grover2016node2vec} on node pairs, and then the best binary operator is selected based on the validation set, while the dot product is used for semi-supervised methods. The results on \textit{CB12} given in Table~\ref{tab:prediction} show the promising results of our proposed graphs. Like \textit{job title classification}, the scores of all network embedding methods, i.e., \textit{N2V}, \textit{GAN}, \textit{M2V} and \textit{HAN} better on $\mathcal{G}^{jj}_{E}$ compared to  $\mathcal{G}^{jj}$, and the heterogeneous models perform best on $\mathcal{G}^{jtj}$. Such results further demonstrate the effectiveness of our proposed method for constructing graphs, whether adding additional information based on tags (i.e., $\mathcal{G}^{jj}_{E}$) or directly adding tags to the graph (i.e., $\mathcal{G}^{jt}$ and $\mathcal{G}^{jtj}$). \textit{BERT} using Hadamard operator performs best, followed by \textit{HAN} on $\mathcal{G}^{jtj}$ with a slight difference of 0.007. However, when we use the dot product used in \textit{HAN} to obtain edge features for \textit{BERT}, the AUC of \textit{BERT} drops sharply to 0.477, while \textit{W2V} only drops a little to 0.763. We will discuss such results in future work. Overall, the results of link prediction also demonstrate the effectiveness of our proposed graphs.

%\vspace{-0.4cm}

\subsubsection{Visualization}
\begin{figure}[th]
    \centering
    \begin{subfigure}[b]{0.22\textwidth}
    \centering
    \includegraphics[width=0.9\textwidth]{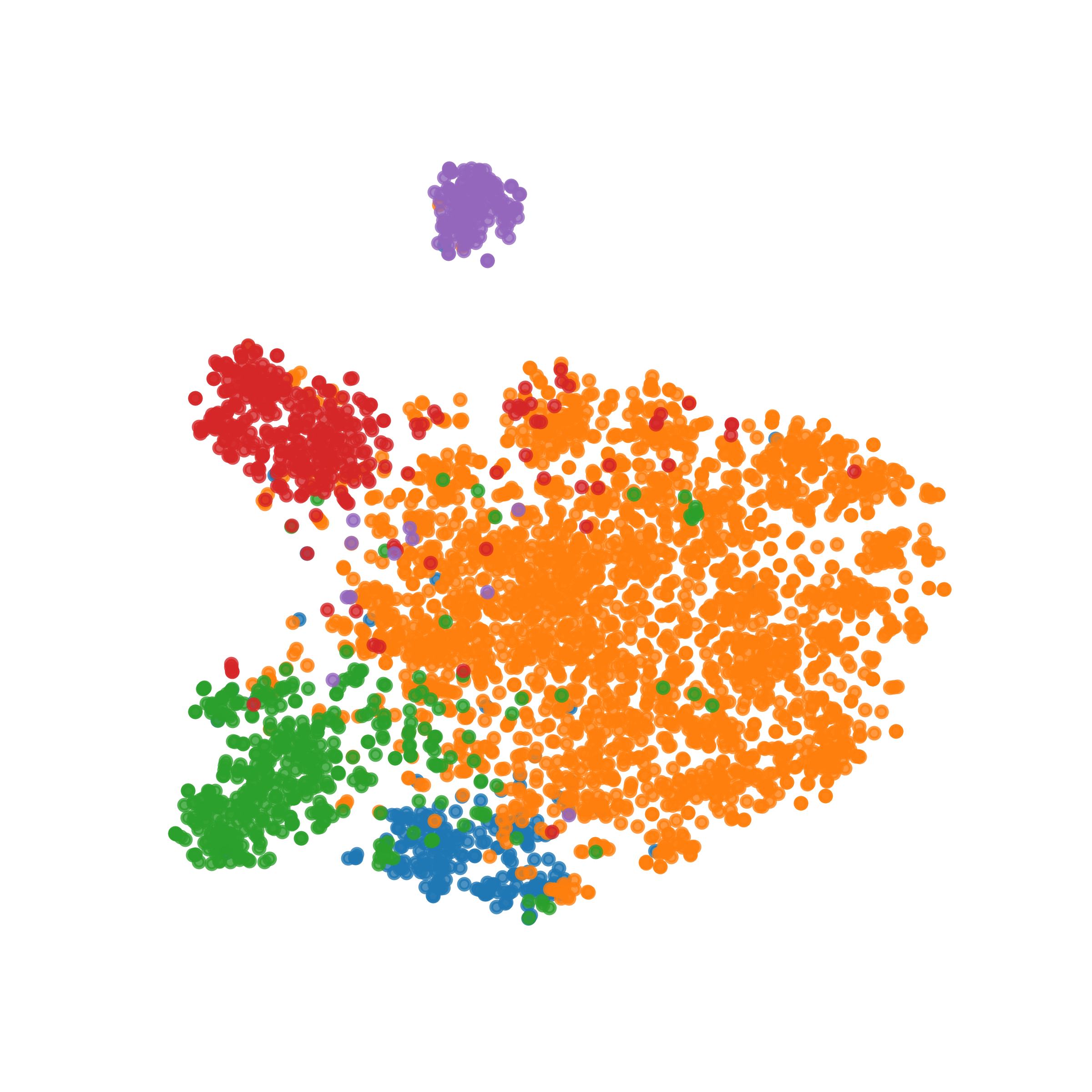}
    \setlength{\abovecaptionskip}{-0.4cm}
    \setlength{\belowcaptionskip}{-0.2cm}
    \caption{HAN ($\mathcal{G}^{jj}$).}
    \label{fig:jj}
    \end{subfigure}
    \hspace{0.1em}
    \begin{subfigure}[b]{0.22\textwidth}
    \centering
    \includegraphics[width=0.9\textwidth]{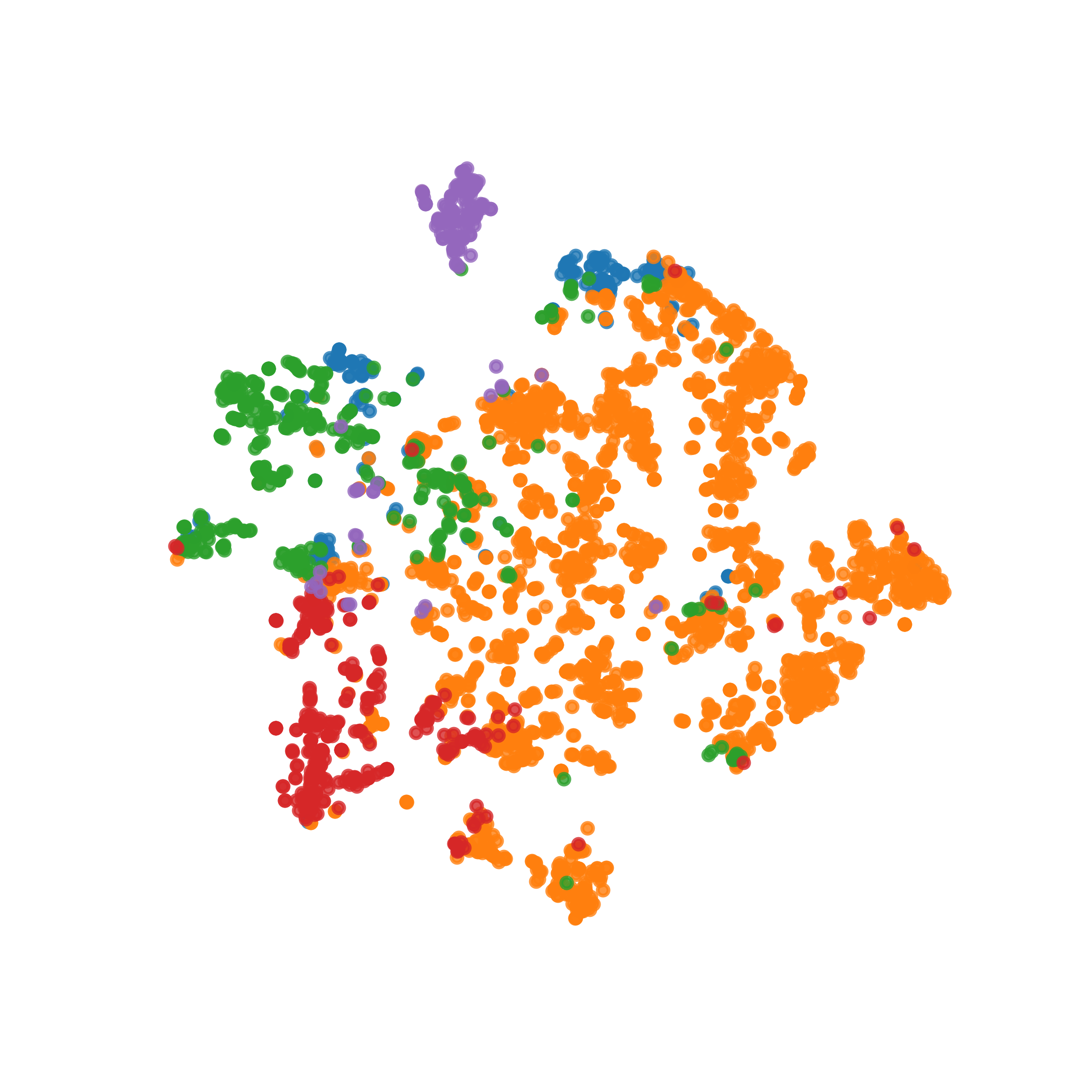}
    \setlength{\abovecaptionskip}{-0.4cm}
    \setlength{\belowcaptionskip}{-0.2cm}
    \caption{HAN ($\mathcal{G}^{jj}_{E}$).}
    \label{fig:jj_E}
    \end{subfigure}
    \hspace{0.1em}
    \begin{subfigure}[b]{0.22\textwidth}
    \centering
    \includegraphics[width=0.9\textwidth]{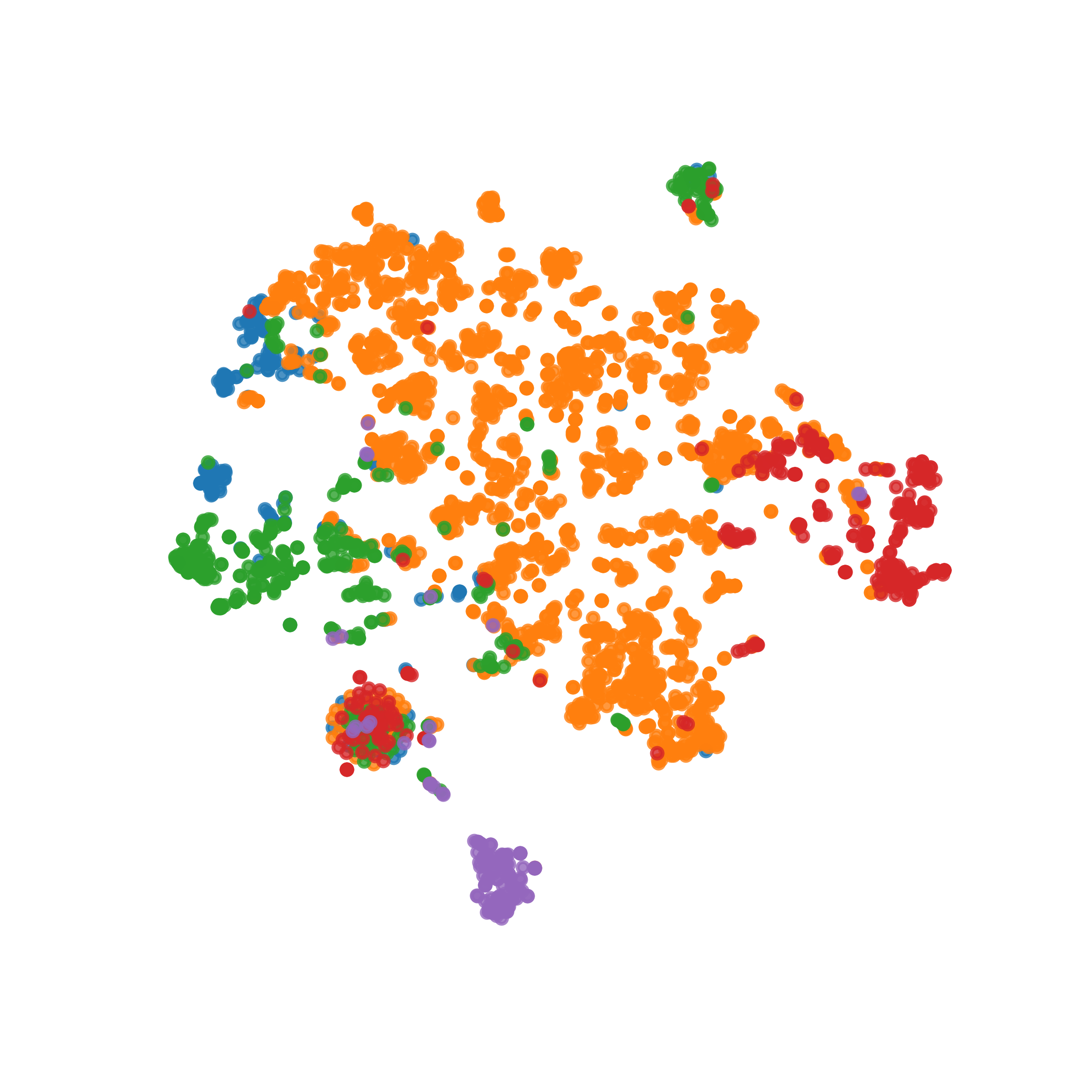}
    \setlength{\abovecaptionskip}{-0.4cm}
    \setlength{\belowcaptionskip}{-0.2cm}
    \caption{HAN ($\mathcal{G}^{jt}$).}
    \label{fig:jt}
    \end{subfigure}
    \hspace{0.1em}
    \begin{subfigure}[b]{0.22\textwidth}
    \centering
    \includegraphics[width=0.9\textwidth]{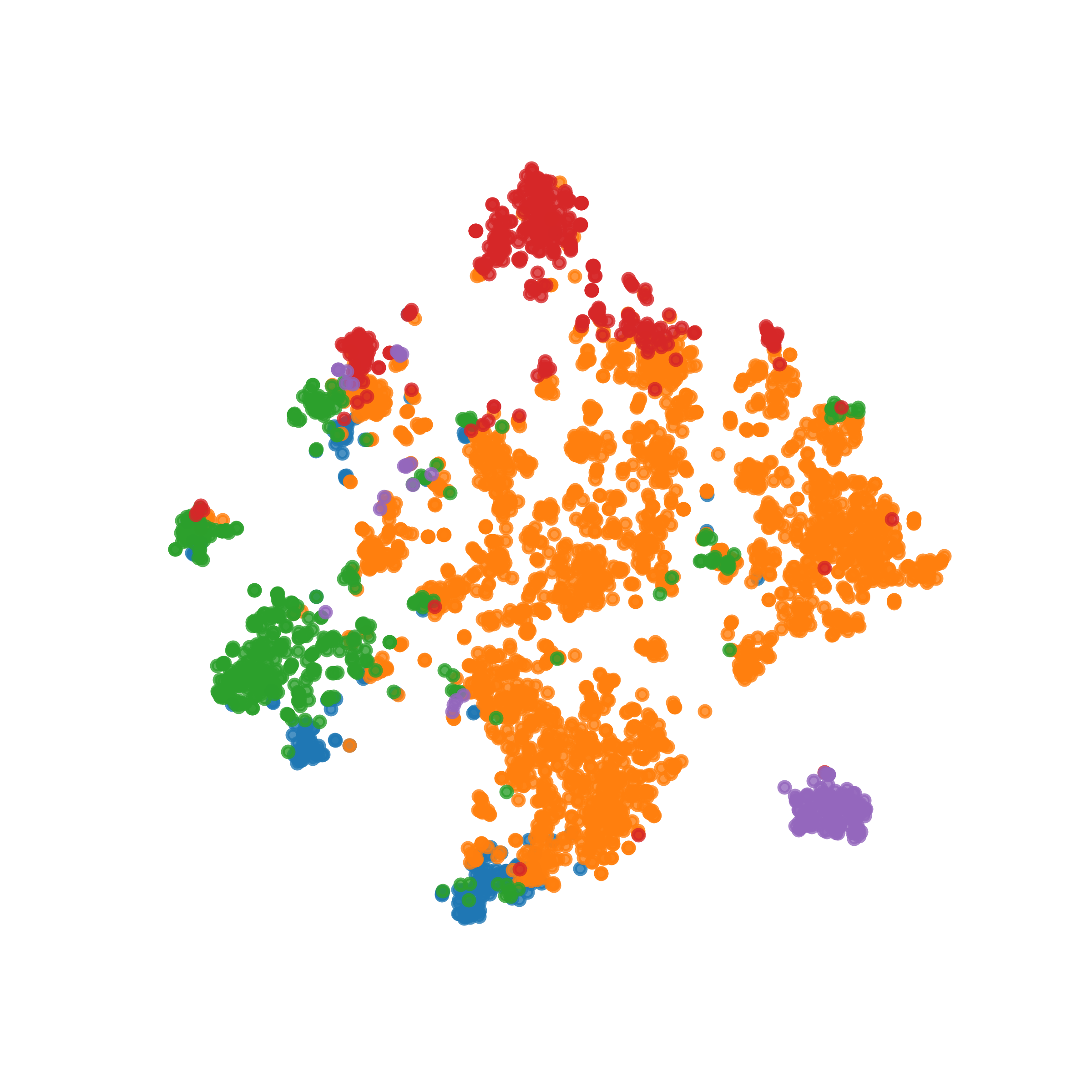}
   \setlength{\abovecaptionskip}{-0.4cm}
    \setlength{\belowcaptionskip}{-0.2cm}
    \caption{HAN ($\mathcal{G}^{jtj}$).}
    \label{fig:jtj}
    \end{subfigure}
%     \begin{subfigure}[b]{0.22\textwidth}
%     \centering
%     \includegraphics[width=1\textwidth]{figs/word2vec.png}
%   \setlength{\abovecaptionskip}{-0.4cm}
%     \setlength{\belowcaptionskip}{-0.2cm}
%     \caption{W2V.}
%     \label{fig:word2vec}
%     \end{subfigure}
%       \begin{subfigure}[b]{0.22\textwidth}
%     \centering
%     \includegraphics[width=1\textwidth]{figs/bert.png}
%   %\setlength{\abovecaptionskip}{-0.4cm}
%     %\setlength{\belowcaptionskip}{-0.2cm}
%     \caption{BERT.}
%     \label{fig:bert}
%     \end{subfigure}
  \caption{Visualization of representations (CB12). \textit{Healthcare support} (green), \textit{Healthcare practitioners and technical} (blue), \textit{Architecture and engineering} (purple),  \textit{Office and administrative support} (orange) and \textit{Transportation and material handling} (red).}
  \label{fig:visualization}
\end{figure}

For a more intuitive comparison, we select five occupations and then visualize the job title representations learned by \textit{HAN} in Figure~\ref{fig:visualization}, with each color corresponding to an occupation category.
%Specifically, we select five occupations ~\footnote{Includes two similar occupations: \textit{Healthcare support} and \textit{Healthcare practitioners and technical}, as well as three non-similar ones: \textit{Architecture and engineering},  \textit{Construction and extraction} and \textit{Transportation and material handling}.} for illustration purposes and use t-SNE ~\citep{van2008visualizing} to reduce the representation dimension.
 Overall, the representations learned by \textit{HAN} on all graphs are clustered into groups. However, when considering tags, representations are easier to be subdivided further in each category. For example, in Figure~\ref{fig:jtj}, the orange occupation can be further divided into three sub-clusters, which proves that adding tag nodes can help capture more detailed information and make the learned representation more informative. This detailed information helps further categorize occupations, as we only classify job titles into the root category (i.e., \textit{MajorGroup}) in this work.

%\vspace{-0.5cm}

\section{Conclusion}
This paper first proposes to enrich \textit{Job-Transition Graph} commonly used in job title representation learning tasks by adding tag-related information or directly adding tag nodes, and then learn job title representations through network embedding methods. This enhancing method can alleviate the sparsity problem in \textit{Job-Transition Graph}, thereby improving the quality of learned representations, as demonstrated in the experimental results of \textit{job title classification} and \textit{job title classification}. In future work, we would like to explore why the Hadamard operator and dot product lead to such different link prediction results for \textit{BERT}. Furthermore, other research lines are (i) considering edge weights when learning from graphs, (ii) classifying job titles into different occupational levels, and (iii) improving the tag generation approach.

\section*{Acknowledgment}
This work is supported by the Randstad research chair in collaboration with MICS Lab, CentraleSupélec, Université Paris-Saclay. We would like to thank the \textit{Mésocentre}~\footnote{\url{http://mesocentre.centralesupelec.fr/}} computing center of CentraleSupélec and École Normale Supérieure Paris-Saclay for providing computing resources.
%We would also like to thank Salim Nibouche for his help with the text preprocessing code.

\bibliography{anthology,custom}
\bibliographystyle{acl_natbib}

\newpage

\appendix

\section{Appendix}
\label{sec:appendix}

\subsection{Job Title Label Assignment}
\label{app: Job Title Label Assignment}
Job titles are not pre-labeled in the original working experience dataset provided by CareerBuilder12. Therefore, for the job title classification task, we use an online third-party API O*Net-SOC AutoCoder~\footnote{\url{http://www.onetsocautocoder.com/plus/onetmatch}} to assign a Standard Occupation Classification code (SOC) 2018 to each job title, as well as a match score (i.e., scores above 70 means that the correct code is accurately predicted at least 70\% of the time). SOC 2018 is a four-level taxonomy structure, including \textit{MajorGroup} (23), \textit{MinorGroup} (98), \textit{BroadGroup} (459) and \textit{DetailedOccupation} (867). For example, \textit{O*Net-SOC AutoCoder} assigns the code 11-2022 (Sales Managers) for the title ``sales director'', which belongs to the level of \textit{DetailedOccupation}. 11-2020 (Marketing and Sales Managers) is \textit{BroadGroup} level,  11-2000 (Advertising, Marketing, Promotions, Public Relations, and Sales Managers) is \textit{MinorGroup} level, and 11-0000 (Management Occupations) is \textit{MajorGroup} level. In this work, we categorize job titles into \textit{MajorGroup}. We have annotated a total of 30,000 job titles. The developer guarantees that the code assigned to the title plus description has an accuracy rate of 85\%. However, only the job title is provided in our experiments, so the SOC 2018 code may be incorrectly assigned. For this reason, we filtered out job titles with scores below 70. Therefore, 12,908 unique job titles remain.

\subsection{Randstad Data Description}
\label{app: Randstad Data Description}
\begin{figure}[th]
  \centering
  \includegraphics[width=\linewidth]{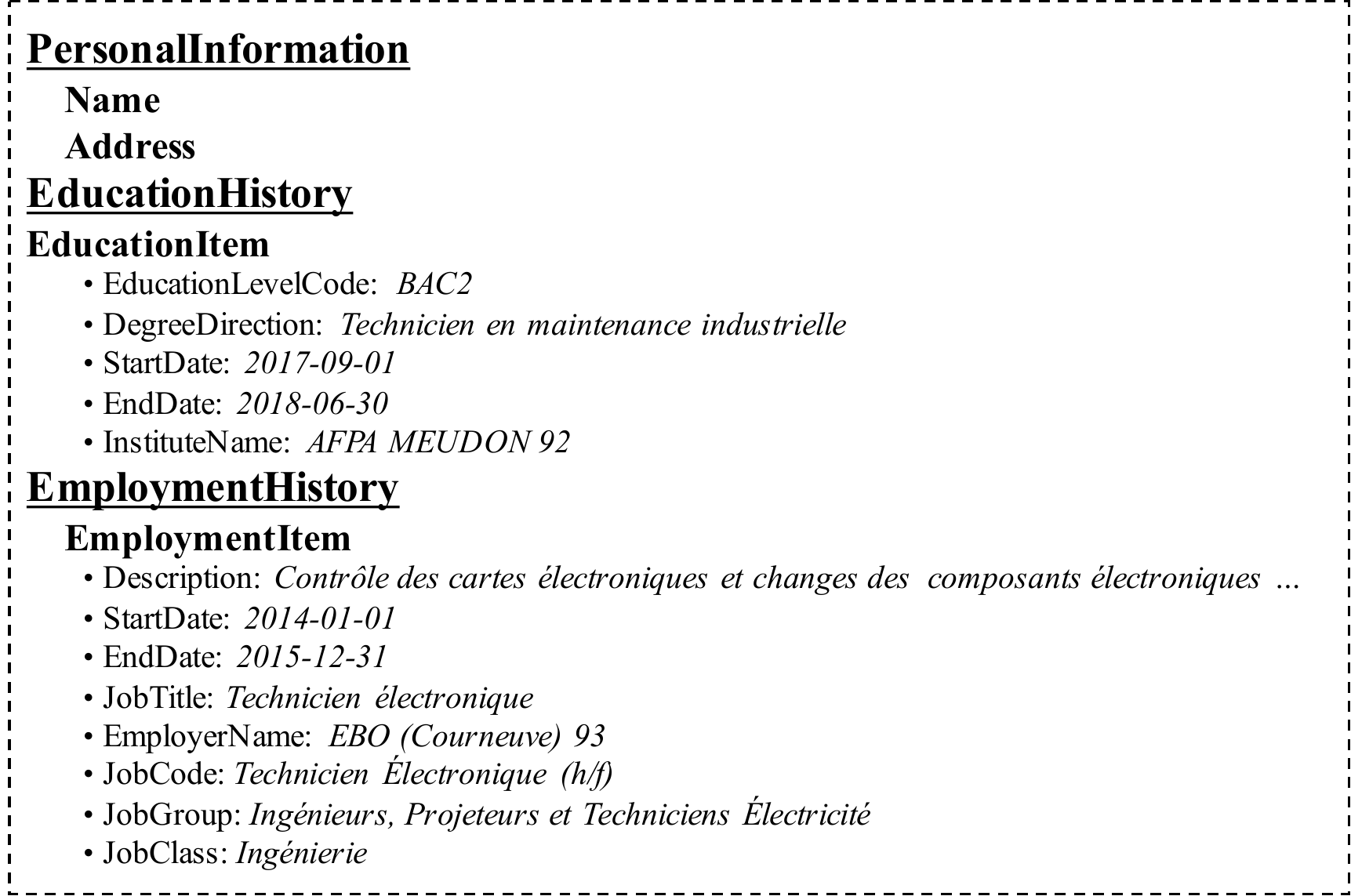}
  \caption{An example of parsed resume in Randstad.}
  \label{fig:parsed_resume}
\end{figure}

Figure~\ref{fig:parsed_resume} shows an example of parsed resume in \textit{Randstad} dataset. We build graphs from \textit{EmploymentHistory}, which contains a \textit{JobTitle}, and its corresponding occupation labels (i.e., \textit{JobCode}, \textit{JobGroup} and \textit{JobClass}). The hierarchical taxonomy structure used in the Randstad dataset has a three-level hierarchy, where \textit{JobCode}s are leaf classes, and each internal class (i.e., \textit{JobGroup}) or root class (i.e., \textit{JobClass}) is the aggregation of all its descendant classes. There are 25 \textit{JobClass}s, 295 \textit{JobGroup}s and 4,443 \textit{JobCode}s, respectively. In this work, we categorize job titles into \textit{JobClass}.

\subsection{Tag Generation}
\label{app: Tags Generation}
\begin{figure}[h]
    \centering
    \begin{subfigure}[b]{0.22\textwidth}
     \centering
     \includegraphics[width=1.18\linewidth]{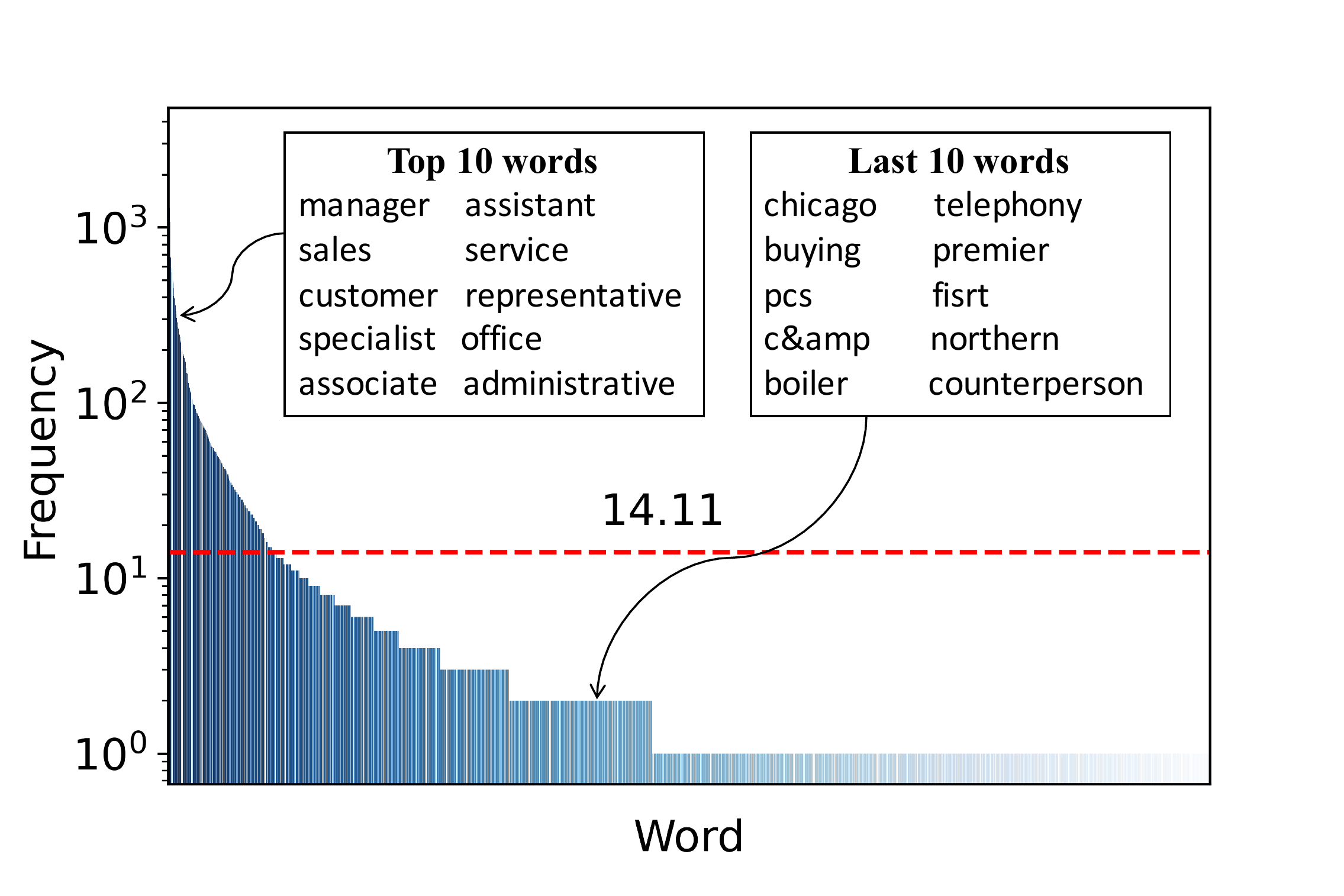}
     \caption{CB12.}
    \end{subfigure}
    \hspace{0.4em}
    \begin{subfigure}[b]{0.22\textwidth}
    \centering
     \includegraphics[width=1.18\linewidth]{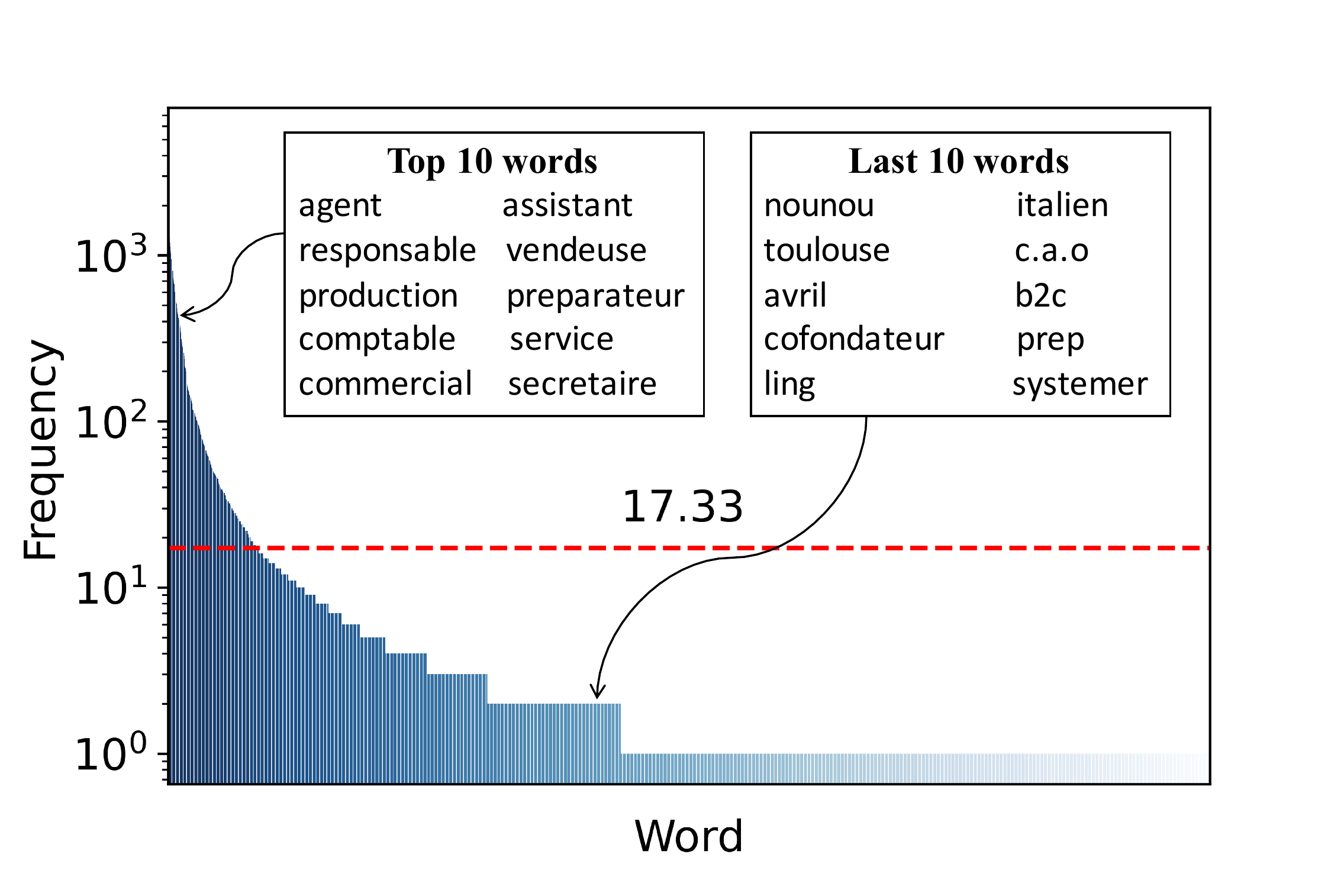}
     \caption{Randstad.}
    \end{subfigure}
   \caption{Word frequency distribution, where the red line represents the average value. \textit{Top-10} words are sorted by frequency, and \textit{Last-10} are randomly selected from the words with a frequency of 2.}
   \label{fig:distribution}
\end{figure}

For both datasets, we first tokenize titles into tokens and remove stopwords, numbers, and punctuation. The word frequency distribution of words in two datasets are shown respectively in Figure~\ref{fig:distribution}, which are subject to the long-tail distribution, similar to the observation in~\citep{zhang2019job2vec}. Most words appear only once, i.e., $53.55\%$ of words only appear one time in \textit{CB12} dataset, and this ratio is $56.55\%$ in \textit{Randstad} dataset. Figure~\ref{fig:distribution} further shows the top ten and last ten frequent words in each dataset. Obviously, high-frequency words like ``manager'' and ``sales'' describe the responsibility or functionality of the job title, while low-frequency words are usually noise or person-specific words. Based on the domain-specific NE tags (i.e., \textit{RES}ponsibility, \textit{FUN}ction) proposed in IPOD~\citep{liu2019ipod}, we then select the Top-200 tokens that appear most frequently and appear in the IPOD NE tag set as tags for each dataset.

% \subsection{Tags Generation}
% \label{app: Tags Generation}

% For both datasets, we use the top tokens with the highest frequencies in job titles as the \textit{Tag} set, and use the open dataset IPOD~\cite{liu2019ipod} to label them as \textit{RES}-tags and \textit{FUN}-tags. It is noted that, since there are no French tags, we directly translate \textit{RES}-tags and \textit{FUN}-tags from English to French. More specific, we first tokenize titles into tokens and remove stopwords, numbers, and punctuation. The word frequency distribution of words in two datasets are shown respectively in Figure~\ref{fig:distribution}, which are subject to the long-tail distribution, similar to the observation in~\citep{zhang2019job2vec}. Most words appear only once, i.e., $53.55\%$ of words only appear one time in \textit{CB12} dataset, and this ratio is $56.55\%$ in \textit{Randstad} dataset. Figure~\ref{fig:distribution} further shows the top ten and last ten frequent words in each dataset. Obviously, high-frequency words like ``manager'' and ``sales'' describe the responsibility or functionality of the job title, while low-frequency words are usually noise or person-specific words. 

\subsection{Baseline Description}
\label{app: Baseline Description}
We explore the following network embedding methods on our proposed graphs to learn job title representation. According to the type of graph, the network embedding methods are naturally divided into \textit{Homogeneous} and \textit{Heterogeneous}. Then, we further categorize each category into \textit{Unsupervised} and \textit{Semi-Supervised} according to whether node labels are provided for learning.

\noindent\textbf{Homogeneous\&Unsupervised}
\begin{itemize}[noitemsep,topsep=2mm,leftmargin=15pt]
    \item \textit{Node2Vec (N2V)}~\citep{grover2016node2vec}: is an extension of DeepWalk with a biased random walk process for neighborhood exploration. % $p = $ and $q = $. 
\end{itemize}

\noindent\textbf{Homogeneous\&Semi-supervised}:
\begin{itemize}[noitemsep,topsep=2mm,leftmargin=18pt]
    \item \textit{GCN}~\citep{kipf2016semi}: is a semi-supervised Graph Neural Network (GNN) that generalizes the convolutional operation to homogeneous graphs.
    \item \textit{GAT}~\citep{velivckovic2017graph}: uses a self-attention strategy to learn the importance between a node and its neighbors.
\end{itemize}

\noindent\textbf{Heterogeneous\&Unsupervised}:
\begin{itemize}[noitemsep,topsep=2mm,leftmargin=15pt]
    \item \textit{Metapath2Vec (M2V)}~\citep{dong2017metapath2vec}: performs meta-path-guided walks and utilizes Skip-Gram to embed heterogeneous graphs.
\end{itemize}

\noindent\textbf{Heterogeneous \& Semi-supervised}:
\begin{itemize}[noitemsep,topsep=2mm,leftmargin=15pt]
    \item \textit{RGCN}~\citep{schlichtkrull2018modeling}: is an extension of \textit{GCN} on heterogeneous graphs, introducing relation-specific transformations based on the type of edges.
    \item \textit{HAN}~\citep{wang2019heterogeneous}: proposes a hierarchical attention mechanism, i.e., node-level and semantic-level for heterogeneous graphs.
\end{itemize}

In addition to the comparison between network embedding methods, we will also compare the representation learned through graphs with the representation obtained by semantic-based methods.

\noindent\textbf{Semantic-based}:
\begin{itemize}[noitemsep,topsep=2mm,leftmargin=15pt]
\item \textit{Word2Vec (W2V)}~\citep{le2014distributed}: the representation of a job title is obtained by averaging word vectors in it. We use word vectors trained on Google News~\footnote{\url{https://code.google.com/archive/p/word2vec/}} for \textit{CB12}, and a pre-trained French embedding model~\citep{fauconnier_2015} for \textit{Randstad}.
\item \textit{BERT}~\citep{devlin2018bert}: the job title representations are obtained by using the bert-as-service package ~\citep{xiao2018bertservice}, a sentence encoding service for mapping variable-length sentences to fixed-length vectors. We default to using the pre-trained BERT models provided by the package, i.e., BERT-Base-Uncased is used for \textit{CB12}, and BERT-Base-Multilingual-Cased (New) for \textit{Randstad}.
\end{itemize}

\subsection{Parameter Settings}
\label{app: Parameter Settings}
Our implementation is based on the PyTorch version of the DGL package ~\citep{wang2019deep}. For \textit{job title classification}, we randomly split the data into training/validation/test sets with a ratio of 60\%/20\%/20\%. We keep the same split ratio on positive/negative edges for \textit{next-job prediction}, where negative edges are randomly picked from unconnected node pairs (i.e., the same size as positive edges). To ensure fairness, we keep the same data split for all methods, and we set the node embedding to 128 for all methods, except for \textit{W2V}.

In the \textit{job title classification} and \textit{next-job prediction} tasks, for unsupervised methods, node representations are learned from the entire dataset. The logistic regression classifier is then trained on both the training and validation sets. Each semi-supervised model is trained on the training set, and the parameters are optimized on the validation set. The final performance is evaluated on the test set. Models are optimized with the Adam~\citep{kingma2014adam} with a learning rate of 1e-3, and we apply $L_{2}$ regularization with value 5e-4. We use an early stop with a patience of 100, i.e., if the validation loss does not decrease in 100 consecutive epochs, we stop training. For models applying the attention mechanism, the dropout rate of attention is set to 0.2. For random-walk-based methods, including \textit{N2V} and \textit{M2V}, we set the window size to 5, walk length to 10, walks per node to 50, the number of negative samples to 5. For \textit{M2V}, we test all meta-paths and report the best performance. For \textit{next-job prediction} task, edge features are represented by applying binary operators~\citep{grover2016node2vec} on pairs of nodes, and then the best operator is chosen based on the validation set, while the dot product is used for all semi-supervised methods. 

We repeat each prediction experiment ten times and report the average performance scores (i.e., Macro-F1 and Micro-F1 for \textit{job title classification} and AUC for \textit{next-job prediction}).

\end{document}